\documentclass[conference]{IEEEtran}
\usepackage{cite}
\usepackage{amsmath,amssymb,amsfonts}
\usepackage{graphicx}
\usepackage{textcomp}
\usepackage{nicefrac}
\usepackage{pifont}
\usepackage{soul}
\usepackage[font=footnotesize,labelfont=bf]{caption}
\usepackage{enumitem}
\usepackage[caption=false, font=footnotesize]{subfig}
\usepackage[table,xcdraw]{xcolor}
\usepackage{titlesec}
\usepackage{anyfontsize}
\usepackage[normalem]{ulem}
\usepackage{wrapfig}
\usepackage{multirow}
\usepackage[export]{adjustbox}
\usepackage{stackengine}    
\usepackage{float}    
\usepackage[edges]{forest}
\usepackage{array}
\usepackage{stfloats}
\usepackage{indentfirst} 
\usepackage{booktabs}
\usepackage{pifont}
\usepackage{makecell}
\usepackage[implicit=false]{hyperref}
\hypersetup{
 colorlinks,
 linkcolor=blue,
 urlbordercolor=black,
 pdfborderstyle={/S/U/W 1}
 }

\definecolor{folderbg}{RGB}{124,166,198}
\definecolor{folderborder}{RGB}{110,144,169}
\definecolor{IGNGREEN}{RGB}{153, 211, 142}
\definecolor{TITLES}{RGB}{153, 211, 142}
\definecolor{TITLES_PRE}{RGB}{247, 212, 188}

\newlength\Size
\titleformat{name=\section}[block]
  {\centering\rmfamily\small}
  {}
  {0pt}
  {\colorsection}
\titlespacing*{\section}{0pt}{\baselineskip}{\baselineskip}
\newcommand{\colorsection}[1]{%
  \colorbox{TITLES_PRE}{\parbox{\dimexpr\linewidth-2\fboxsep}{\thesection\ #1}}}

\def\BibTeX{{\rm B\kern-.05em{\sc i\kern-.025em b}\kern-.08em
    T\kern-.1667em\lower.7ex\hbox{E}\kern-.125emX}}
    
\tikzset{%
  folder/.pic={%
    \filldraw [draw=folderborder, top color=folderbg!50, bottom color=folderbg] (-1.05*\Size,0.2\Size+5pt) rectangle ++(.75*\Size,-0.2\Size-5pt);
    \filldraw [draw=folderborder, top color=folderbg!50, bottom color=folderbg] (-1.15*\Size,-\Size) rectangle (1.15*\Size,\Size);},
  file/.pic={%
    \filldraw [draw=folderborder, top color=folderbg!5, bottom color=folderbg!10] (-\Size,.4*\Size+5pt) coordinate (a) |- (\Size,-1.2*\Size) coordinate (b) -- ++(0,1.6*\Size) coordinate (c) -- ++(-5pt,5pt) coordinate (d) -- cycle (d) |- (c) ;},
}

\forestset{%
  declare autowrapped toks={pic me}{},
  pic dir tree/.style={%
    for tree={folder,font=\ttfamily,grow'=0,},
    before typesetting nodes={%
      for tree={edge label+/.option={pic me},},},
  },
  pic me set/.code n args=2{%
    \forestset{%
      #1/.style={inner xsep=2\Size,pic me={pic {#2}},}
    }
  },
  pic me set={directory}{folder},
  pic me set={file}{file},
}

\newenvironment{Tabular}[2][1]
  {\def\arraystretch{#1}\tabular{#2}}
  {\endtabular}

\begin{document}

\twocolumn[{
  \begin{@twocolumnfalse}
    \centering\underline{\makebox[16cm]{\LARGE{FLAIR: French Land cover from Aerospace ImageRy.}}} \\
    \vspace{-0.38cm}
    \begin{figure}[H]
    \centering
    \includegraphics[width=2.052\linewidth]{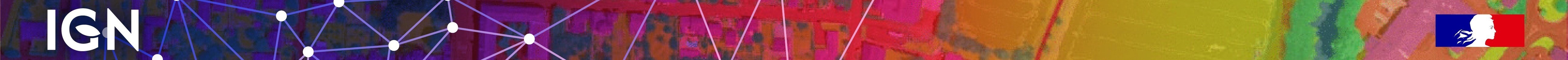}
    \end{figure}
    \centering
    \vspace{-0.3cm}
    \Large{\textit{Challenge \mbox{FLAIR \#2}: }} \\
    \Large{\textit{textural and temporal information for semantic segmentation\\from multi-source optical imagery}} \\
    \vspace{+0.5cm}
    \normalsize{Anatol Garioud, Apolline De Wit, Marc Poupée, Marion Valette, Sébastien Giordano, Boris Wattrelos}    \\
    \vspace{+0.1cm}
    \normalsize{Institut national de l’information géographique et forestière (IGN), France} \\
    \vspace{+0.1cm}
    \centering\normalsize{\textit{ai-challenge@ign.fr}}                  
    \vspace{-0.2cm}
    \begin{figure}[H]
    \centering
    {\setlength{\fboxsep}{0pt}%
    \setlength{\fboxrule}{1.5pt}%
    \fbox{\includegraphics[width=2.04\linewidth]{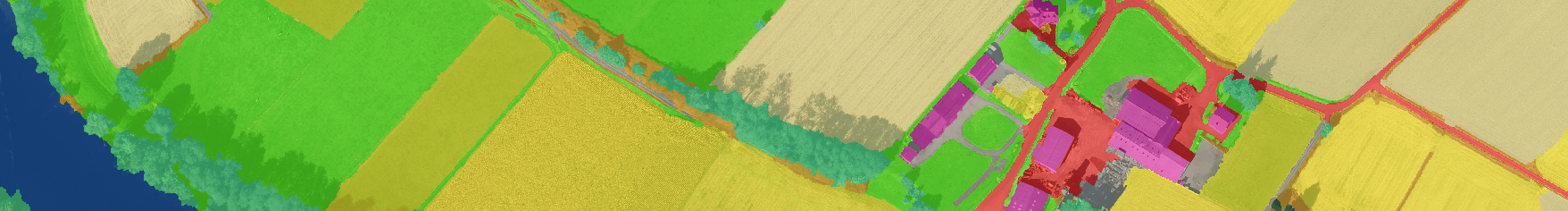}}}
    \end{figure}    
  \end{@twocolumnfalse}
}]

\section*{\textbf{Dataset overview}}
\vspace{-0.5cm}
\begin{table}[h!]
\centering
\renewcommand{\arraystretch}{1.5}
\begin{Tabular}[1.5]{|p{8.2cm}|}
\hline \rowcolor[HTML]{C0C0C0} \textbf{Figures}                    \\  \hline
\end{Tabular}
\par\vskip1.2pt
\begin{Tabular}[0.8]{|p{8.2cm}|}
\hline
\rowcolor[HTML]{dbdbdb} \color{black}\ding{212} \color{black} 20,384,841,728 pixels annotated at 0.20\:m spatial resolution \\
\rowcolor[HTML]{dbdbdb} \color{black}\ding{212} \color{black} 77,762 patches (512$\times$512) \\
\rowcolor[HTML]{dbdbdb} \color{black}\ding{212} \color{black} 51,244 satellite acquisitions with broader spatial context \\
\rowcolor[HTML]{dbdbdb} \color{black}\ding{212} \color{black} 50 spatio-temporal domains and 916 areas covering 817 km²\\
\rowcolor[HTML]{dbdbdb} \color{black}\ding{212} \color{black} 13 semantic classes (+6 optional ones)        \\[1em] \hline
\end{Tabular}
\end{table}

\vspace{-0.5cm}

\begin{table}[h!]
\centering
\renewcommand{\arraystretch}{1.5}
\begin{Tabular}[1.5]{|p{8.2cm}|}
\hline \rowcolor[HTML]{C0C0C0} \textbf{Structure}                    \\  \hline
\end{Tabular}
\par\vskip1.2pt
\hspace{0.02cm}
\begin{Tabular}[0.8]{|p{8.2cm}|}
\hline 
\rowcolor[HTML]{dbdbdb}  {\footnotesize
\begin{forest}
  pic dir tree, where level=0{}{directory,},
  for tree={ s sep=0.1cm, l sep=1cm, font=\rmfamily }
  [\textbf{Dataset}
    [\textbf{aerial train\textbackslash val}
      [domain \_ year
        [area 
          [img
            [IMG\_ID.tif, file]]           
            ]
      ]
    ]
    [\textbf{sen train\textbackslash val}
      [domain \_ year
        [area 
          [sen
            [SEN2\_data.npy, file]
            [SEN2\_masks.npy, file]
            [SEN2\_products.txt, file]]           
            ]
      ]
    ]
    [\textbf{labels train}
      [domain\_year
        [area
          [msk
            [MSK\_ID.tif, file]]    
        ]  
      ]
    ]
    [metadata\_aerial.json, file]
    [centroids\_sp\_to\_patch.json, file]
  ]
\end{forest}}      \\ \hline
\end{Tabular}
\vspace{-4mm}
\end{table}

\section*{\textbf{Context}}
According to a report by the Food and Agriculture Organization of the United Nations (FAO) in 2015 \cite{FAO}, a significant portion of the world's soil resources are in a condition that can be classified as fair, poor, or very poor. This degradation of soils, coupled with the loss of biodiversity, has far-reaching implications for the state of ecosystems and their long-term sustainability. Soils play a vital role in providing a range of ecosystem services. They serve as natural habitats for numerous plant and animal species, act as a crucial carbon sink by absorbing CO$_{2}$ (to the extent that they are the largest carbon sink, surpassing the atmosphere and all vegetation and animals on Earth's surface), filter rainwater, support food production, and function as the planet's largest water reservoir. The degradation of soils and biodiversity can be attributed in large part to the process of land artificialization, with urban sprawl being a significant contributing factor. This growing phenomenon has raised concerns among public authorities, who recognize the importance of monitoring the state of territories. Artificialization is defined as the long-term deterioration of the ecological functions of soil, including its biological, hydrological, climatic, and agronomic functions, resulting from its occupation or use \cite{LoiClimat}.\\

The French National Institute of Geographical and Forest Information (IGN)\cite{IGN}, in response to the growing availability of high-quality Earth Observation (EO) data, is actively exploring innovative strategies to integrate these data with heterogeneous characteristics. As part of their initiatives, the institute employs artificial intelligence (AI) tools to monitor land cover across the territory of France and provides reliable and up-to-date geographical reference datasets.\\ 

The \mbox{FLAIR \#1} dataset, which focused on aerial imagery for semantic segmentation, was released to facilitate research in the field. Building upon this datset, the \mbox{FLAIR \#2} dataset extends the capabilities by incorporating a new input modality, namely Sentinel-2 satellite image time series, and introduces a new test dataset Both \mbox{FLAIR \#1} and \#2 datasets are part of the currently explored or exploited resources by IGN to produce the French national land cover map reference \textit{Occupation du sol à grande échelle} (OCS-GE).

\section*{\textbf{Multi-modality fusion challenge}}
The growing importance of EO in the monitoring and understanding of Earth's physical processes, and the diversity of data now publicly available naturally favours multi-modal approaches that take advantage of the distinct strengths of this data pool. Remote sensing data have several main characteristics that are of crucial importance depending on the intended purpose. Spatial, temporal and spectral resolutions will influence the choice of data and their importance in a process. The complexity of integrating these different data tend to promotes the use of machine learning for their exploitation.\\

This \mbox{FLAIR \#2} challenge organized by IGN proposes the development of multi-resolution, multi-sensor and multi-temporal aerospace data fusion methods, exploiting deep learning computer vision techniques.\\

The \mbox{FLAIR \#2} dataset hereby presented includes two very distinct types of data, which are exploited for a semantic segmentation task aimed at mapping land cover. The data fusion workflow proposes the exploitation of the fine spatial and textural information of very high spatial resolution (VHR) mono-temporal aerial imagery and the temporal and spectral richness of high spatial resolution (HR) time series of Copernicus Sentinel-2 \cite{Sentinel-2-paper} satellite images, one of the most prominent EO mission. Although less spatially detailed, the information contained in satellite time series can be helpful in improving the inter-class distinction by analyzing their temporal profile and different responses in parts of the electromagnetic (EM) spectrum. \\

\titleformat{name=\section}[block]
  {\centering\rmfamily\small}
  {}
  {0pt}
  {\colorsection}
\titlespacing*{\section}{0pt}{\baselineskip}{\baselineskip}
\renewcommand{\colorsection}[1]{%
  \colorbox{TITLES}{\parbox{\dimexpr\linewidth-2\fboxsep}{\thesection\ #1}}}

\section*{\textbf{Spatial and temporal domains definition}}

\textbf{Spatial domains and divisions}: as for the \mbox{FLAIR \#1} dataset, a spatial domain is equivalent to a French 'département' which is a french sub-region administrative division. While the spatial domains can be geographically close, heavy pre-processing of the radiometry of aerial images independently per 'département' create important differences (see \cite{FLAIR-1}). Each domain has a varying number of areas subdivided in patches of same size across the dataset. \\

\begin{figure}[!htpb]
    \centering
    \includegraphics[width=1\linewidth]{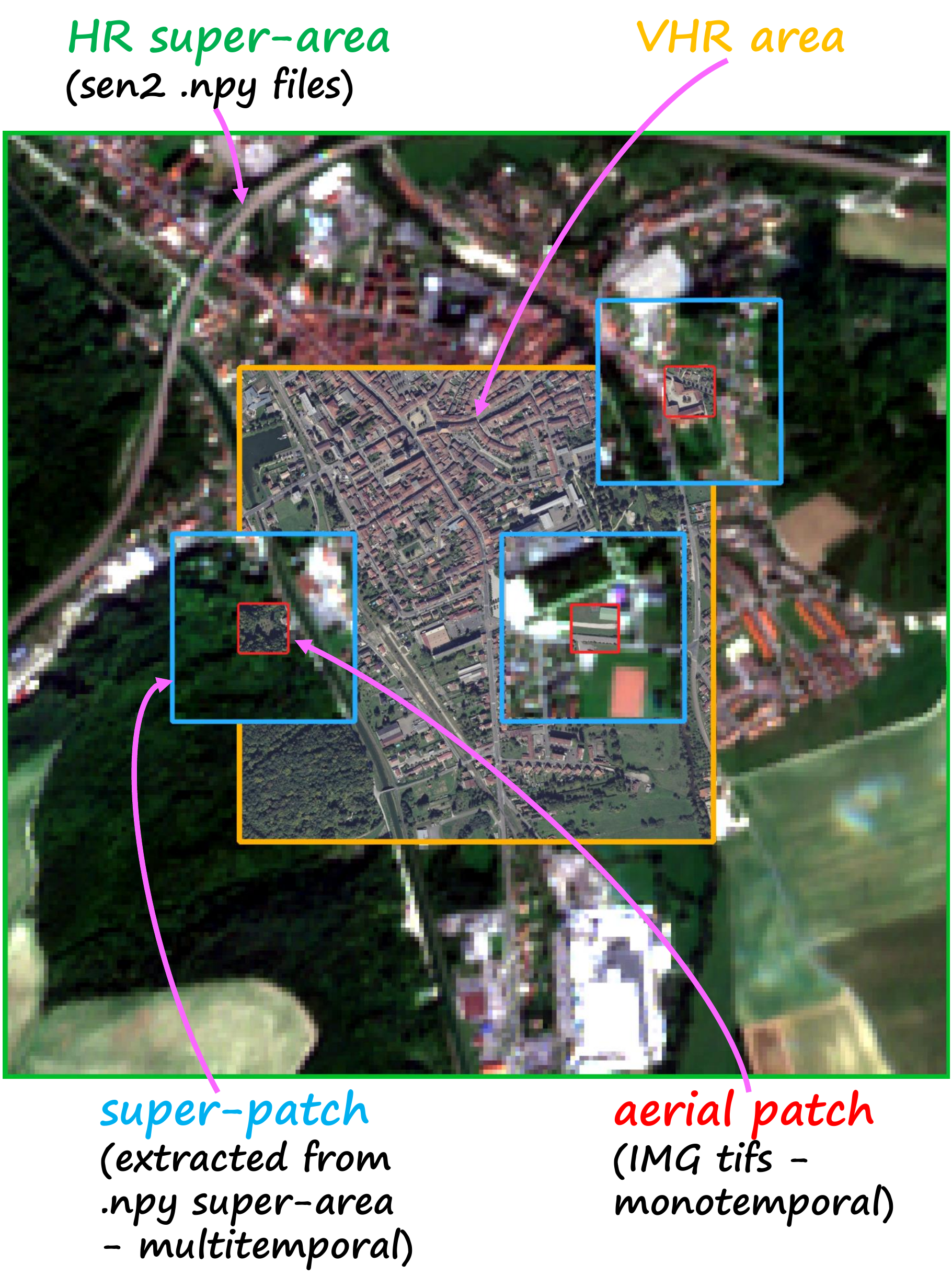}
    \caption{Spatial definitions of the FLAIR~\#2 dataset: HR Sentinel-2 super-area, VHR aerial area, HR Sentinel-2 super-patch and VHR aerial patch.}
    \label{fig:areas}
\end{figure}

While these areas were initially defined to contain sufficient spatial context by taking into account aerial imagery, the strong difference in spatial resolution with satellite data means that they consist of few Sentinel-2 pixels. Therefore, in order to also provide a minimum of context from the satellite data, a buffer was applied to create \textit{super-areas}. This allows, for every patch of the dataset to be associated to a \textit{super-patch} of Sentinel-2 data with sufficient size through a large footprint. Figure~\ref{fig:areas} illustrates the different spatial units of the dataset. \\

\begin{figure}[!htpb]
    \centering
    \includegraphics[width=1\linewidth]{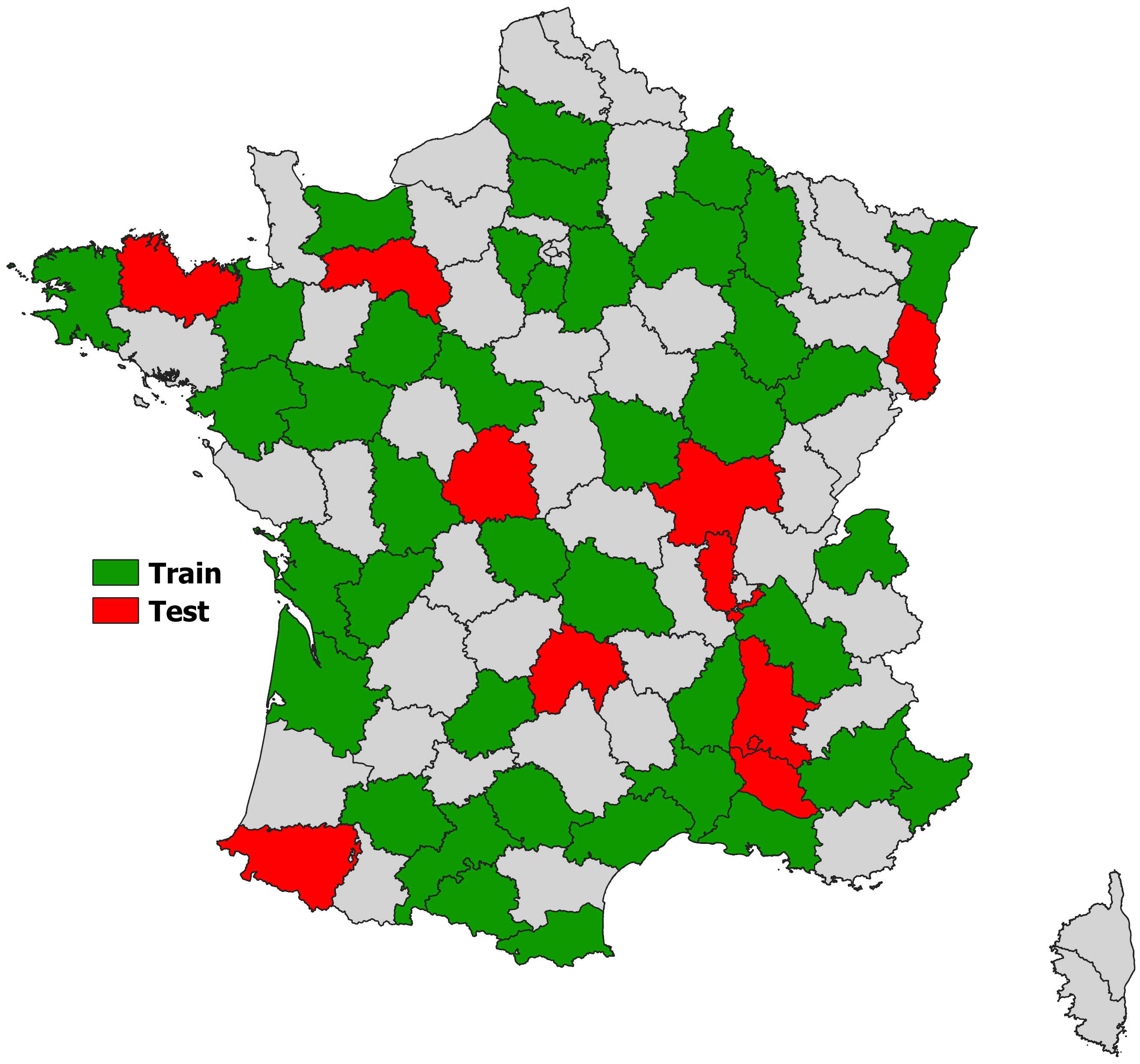}
    \caption{The 50 spatial domains in France of the \mbox{FLAIR \#2} dataset and the train/test split.}
    \label{fig:train_test}
\end{figure}

\textbf{Temporal domains}: they are twofold, on the one hand the date of acquisition of the aerial imagery (which varies in terms of year, month, days) and on the other hand by the satellite acquisitions, varying in terms of months and days.\\

\textbf{Dataset extent}: The dataset includes 50 spatial domains (Figure~\ref{fig:train_test}) representing the different landscapes and climates of metropolitan France. The train dataset constitute 4/5 of the spatial domains (40) while the remaining 1/5 domains (10) are kept for testing. This test dataset introduces new domains compared to the \mbox{FLAIR \#1} test dataset. Some domain are in common but areas within those domains are distinct. The \mbox{FLAIR \#2} dataset covers approximately 817 km ² of the French metropolitan territory.\\

\begin{figure*}[!hb]
    \centering
    \includegraphics[width=1\linewidth]{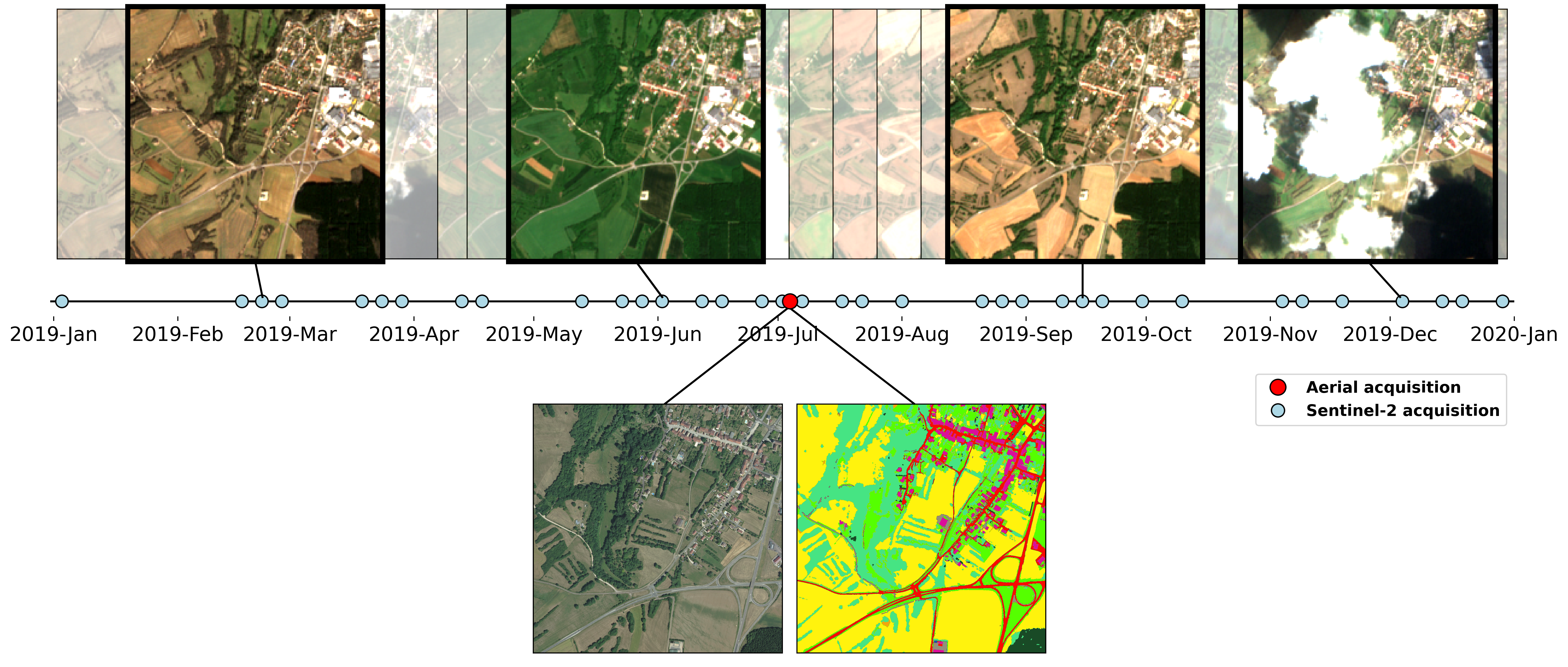}
    \caption{Example of mono-temporal aerial acquisition and annotations over an area (\textit{bottom}) and Sentinel-2 time series acquisitions of the corresponding super-area (\textit{top}) with four acquisitions examples.}
    \label{fig:frise}
\end{figure*}

\section*{\textbf{Dataset resolutions}}

We hereby define different resolutions of the data used in the \mbox{FLAIR \#2} dataset.\\

\textbf{Spatial resolution}: spatial resolution of aerial images can slightly vary depending on the camera used (refer to \cite{FLAIR-1} for more information), but all images are resampled to a 0.2\:m spatial resolution. In contrast, the Sentinel-2 MultiSpectral Instrument (MSI) sensor acquires images at 10, 20 and 60\:m spatial resolutions. The 60\:m bands mainly intended for atmospheric corrections are not taken into account.\\   

\textbf{Spectral resolution}: aerial images acquire 4 spectral bands, namely blue, green, red and near-infrared \cite{FLAIR-1}. Satellite images have more spectral depth with 10 bands, ranging from the visible to the medium infrared parts of the EM spectrum. Sentinel-2 original bands are described in Table~\ref{table:bands_sentinel2}.\\

\textbf{Temporal resolution}: aerial images are usually acquired between the months of April and November and it takes three years to cover the entire French territory (see temporal domains in \cite{FLAIR-1}). Therefore, over an area, a single aerial image is available in the \mbox{FLAIR \#2} dataset. While orbiting, the Sentinel-2 constellation composed of two satellites (Sentinel-2 A\&B) has a revisit frequency of 5 days at the equator, less as you move towards the poles. In the case of the \mbox{FLAIR \#2} dataset, for each area of the dataset, all images acquired by Sentinel during the same acquisition year are considered, including cloudy dates. Nonetheless, for each area, the number of acquisitions varies because of relative orbits and orbit overlaps. An example of aerial and satellite acquisition over an area of the dataset is illustrated in Figure~\ref{fig:frise}.\\

\begin{table}[t]
\footnotesize
\centering
\renewcommand{\arraystretch}{1}
\begin{tabular}{c|c|c|c}
\makecell{Band\\ \:} & \makecell{Central wavelength\\(nm)} & \makecell{Bandwidth\\(nm)} & \makecell{Spatial resolution\\(m)} \\\hline
2 & 490 & 65 & 10 \\
3 & 560 & 35 & 10 \\ 
4 & 665 & 30 & 10 \\ 
5 & 705 & 15 & 20 \\ 
6 & 740 & 15 & 20 \\ 
7 & 783 & 20 & 20 \\ 
8 & 842 & 115 & 10 \\
8a & 865 & 20 & 20 \\
11 & 1610 & 90 & 20 \\ 
12 & 2190 & 180 & 20 \\ \hline
\end{tabular}
\caption{Original spatial and spectral resolutions of Sentinel-2 images.}
\label{table:bands_sentinel2}
\end{table}

In summary, both aerial and satellite data-types included in the \mbox{FLAIR \#2} dataset have strong heterogeneity in terms of spatial (factor of 50), spectral (4 bands versus 10) and temporal (mono-temporal versus time series) resolutions. This opens up challenging perspectives for data fusion schemes in both technical and thematic aspects.

\section*{\textbf{Data sources and pre-processing}}
For details about aerial images (ORTHO HR\textsuperscript{\textregistered}) and associated elevation data, as well as pre-processing, refer to the \mbox{FLAIR \#1} datapaper \cite{FLAIR-1}.\\

Technical details about Sentinel-2 can be found in \cite{Sentinel-2-paper}. The images were downloaded from the Sinergise API \cite{SenHub}  as Level-2A products (L2A) which are atmospherically corrected using the Sen2Cor algorithm \cite{Sen2Cor}. L2A products provide Bottom-Of-the-Atmosphere (BOA) reflectances, corresponding to a percentage of the energy the surface reflects. L2A products also deliver pixel-based cloud (CLD) and snow (SNW) masks at 20\:m spatial resolution. Sentinel-2 images are typically provided as 110$\times$110\:km (with 10\:km overlay) squared ortho-image in UTM/WGS84 projection. However, in order to limit the size of the data and due to the wide extent of the dataset, only the super-areas were downloaded. Concerning Sentinel-2 pre-processing, the 20\:m spatial resolution bands are first resampled during data retrieval to 10\:m by the nearest interpolation method. Same approach is adopted for the cloud and snow masks. Due to the relative orbits of Sentinel-2 some images contain nodata pixels (reflectances at 0). As all Sentinel-2 images during the aerial image acquisition year are gathered all dates containing such nodata were removed. It must be remarked that the length of time series and the acquisition dates thus varies for each super-area. Table~\ref{table:s2-ts} provides information about the number of dates included in the filtered Sentinel-2 time series for the train and test datasets. In average, each area is acquired on 55 dates over the course of a year by the satellite imagery. 

\begin{table}[!htpb]
\centering
\renewcommand{\arraystretch}{1.5}
\begin{tabular}{lcccc}
& \multicolumn{3}{c}{acquisitions per super-area} & \\\cline{2-4}
\makecell[l]{Sentinel-2 time series (1 year)} & min & max & mean & total \\ \hline
\multicolumn{1}{l|}{train dataset} & 20 & 100 & \multicolumn{1}{c|}{55} & 757 \\
\multicolumn{1}{l|}{test dataset} & 20 & 114 & \multicolumn{1}{c|}{55} & 193 \\ \hline
\end{tabular}
\caption{Number of acquisitions (dates) in the Sentinel-2 times series of one year (corresponding to the year of aerial imagery acquisition).}
\label{table:s2-ts}
\end{table}

Note that cloudy dates are not suppressed from the time series. Instead, the masks are provided and can be used to filter the cloudy dates if needed. The resulting Sentinel-2 time series are subsequently reprojected into the Lambert-93 projection (EPSG:2154) which is the one of the aerial imagery.

\section*{\textbf{Data description, naming conventions and usage}}
The \mbox{FLAIR \#2} dataset is composed of 77,762 aerial imagery patches, each 512$\times$512 pixels, along with corresponding annotations, resulting in a total of over 20 billion pixels. The patches correspond to 916 areas  distributed across 50 domains and cover approximately 817\:km$^{2}$. The area sizes and the number of patches per area vary but are always a multiple of 512 pixels at a resolution of 0.20\:meters. Additionally, the dataset includes 55,244 satellite super-areas acquisitions that have a buffer of 5 aerial patches (512\:m) surrounding each aerial area. Description of the data is provided bellow: \\ 

\begin{figure*}[!ht]
    \centering
    \includegraphics[width=0.9\linewidth]{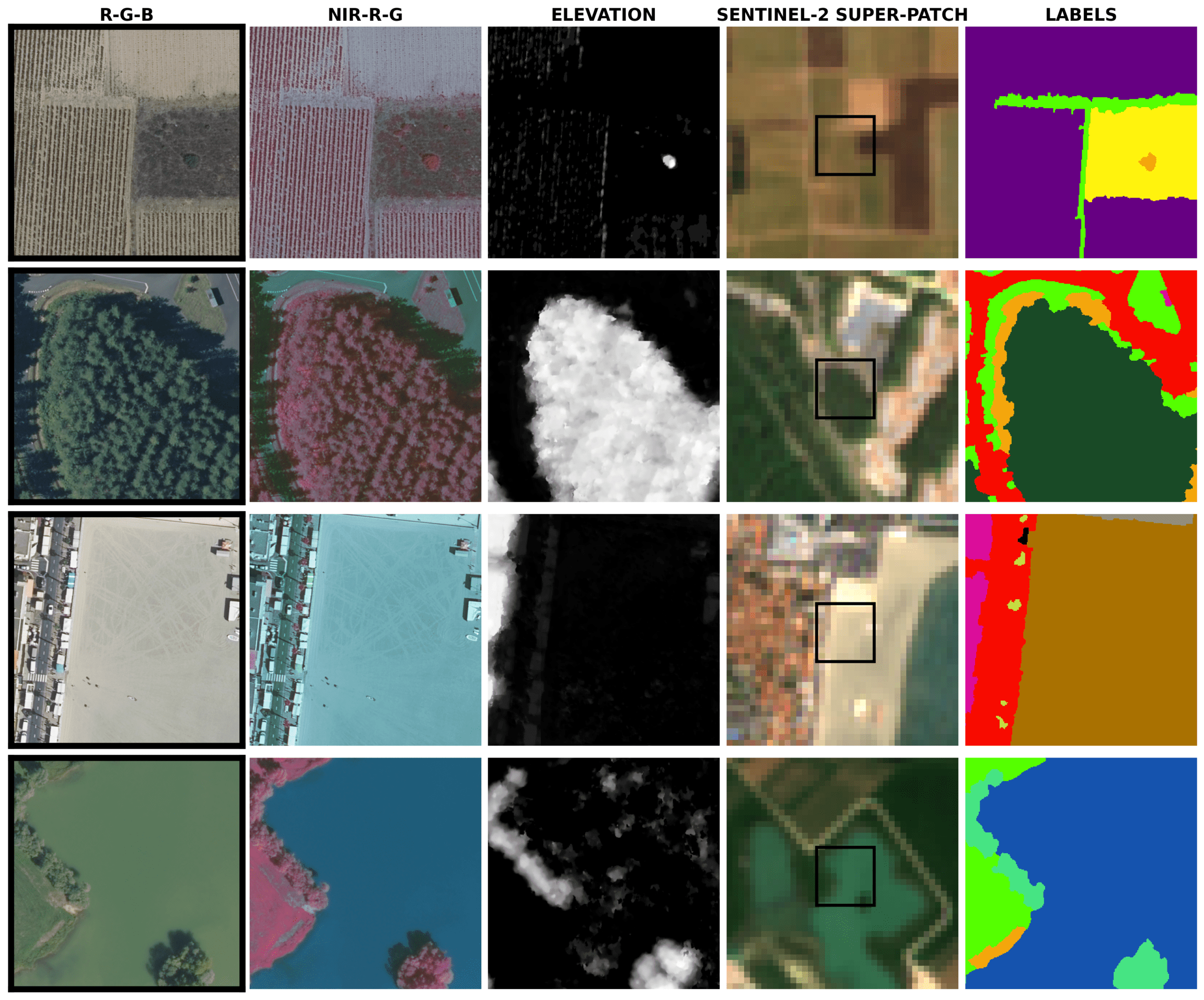}
    \caption{Example of input and supervision data: true color composition, near-infrared color composition, elevation band, Sentinel-2 true color composition super-patch and supervision masks. The data from the first three columns are retrieved from the IMG files, the super-patch from SEN numpy files while the last column corresponds to the MSK files.}
    \label{fig:patches}
\end{figure*}

\begin{itemize}
    \setlength\itemsep{1.2em}
    
    \item[$\blacktriangleright$] The \textbf{aerial input patches (IMG)} consist of 5 channels, similar to the \mbox{FLAIR \#1} dataset. These channels include blue, green, red, near-infrared, and elevation bands, all encoded as 8-bit unsigned integer datatype. The aerial patches are named as \textit{IMG\_ID}, with a unique identifier (ID) across the dataset assigned to each patch.

    \item[] A file named \textbf{\textit{flair\_aerial\_metadata.json}} contains metadata for each of the aerial patches. This JSON file provides detailed information such as the date and time of acquisition, the geographical location of the patch centroid (x, y), the mean altitude of the patch (z), and the type of camera used. For more in-depth descriptions of these metadata attributes, please refer to the documentation provided in \cite{FLAIR-1}.
    
    \item[$\blacktriangleright$] The \textbf{Sentinel-2 super-areas (SEN2) data} is composed of several elements - \textit{data}, \textit{masks}, \textit{products} and a \textit{JSON} file to match aerial and satellite imagery - :\\
    \begin{itemize}
        \setlength\itemsep{0.5em}
        \item the super-area reflectance time series is stored in the \textit{\textbf{SEN2\_xxxx\_data.npy}} files. These files contain 4D NumPy arrays with a shape of \textit{T}$\times$\textit{C}$\times$\textit{H}$\times$\textit{W}, where \textit{T} represents the acquisition dates (which can vary for each file), \textit{C} represents the 10 spectral bands of Sentinel-2, and \textit{H} and \textit{W} denote the height and width dimensions of the data, respectively. The data is stored as uint16 datatype, which differs from the acquisition datatype mentioned in the SenHub reference provided \cite{SenHub}. It's important to note that the data in these files is provided without any additional processing or modifications.
    
        \item the super-area cloud and snow masks are stored in the \textit{\textbf{SEN2\_xxxx\_masks.npy}} files. These files have a similar shape as the data files, with a 4D array format of \textit{T}$\times$\textit{C}$\times$\textit{H}$\times$\textit{W}. However, they consist of only two channels, representing the snow masks and cloud masks, respectively, in that order. The values in the masks range from 0 to 100 and indicate the probability of cloud or snow occurrence for each pixel. A value of 100 indicates a high probability.

        \item the names of the Sentinel-2 time series products are listed in the \textit{\textbf{SEN2\_xxxx\_products.txt}} file. This file provides additional information for each acquisition, including the Sentinel-2 platform (S2A or S2B), the acquisition date (which corresponds to the first date mentioned in the product name), the acquisition time, the orbit number and tile name associated with the product. These details help identify and differentiate the specific products within the Sentinel-2 time series dataset.
    \end{itemize}
    \medskip

    Additionally, \textbf{\textit{flair-2\_centroids\_sp\_to\_patch.json}} file is provided alongside the data. This file plays a role in dynamically cropping the satellite super-areas into super-patches during the data loading process. The JSON file uses the aerial patch name (\textit{e.g.}, IMG\_077413) as the key and provides a list of two indexes (\textit{e.g.}, [13,25]) that represent the data-coordinates of the aerial patch centroids. Using these coordinates and a specified number of pixels (referred to as \textit{sat\_superpatch\_size}), super-patches are extracted from the satellite data. For the experiments, the default \textit{sat\_superpatch\_size} is set to 40, resulting in super-patches with a spatial size of 40*40 pixels. This size corresponds approximately to two aerial patches on each side of the centroid.\\

    The pattern \textbf{\textit{xxxx}} in the file names corresponds to the format domain\_year-areanumber\_arealandcoverletters (\textit{e.g.}, D077\_2021-Z9\_AF). The \textit{arealandcoverletters} represent the two broad types of land cover present in the area. For more detailed information about the specific land cover types, please refer to \cite{FLAIR-1}.
    
    \item[$\blacktriangleright$] The \textbf{annotation patches (MSK)} consist of a single channel with values ranging from 1 to 19, encoded as an 8-bit unsigned integer datatype. These files are named as \textit{MSK\_ID}, where ID corresponds to the same identifier used for the aerial imagery patches.\\
    
    It is important to note that annotations are limited to the boundaries of aerial imagery areas and do not extend to satellite super-areas. In addition, annotations derived from aerial imagery correspond to the specific date the images were captured. However, certain evolving classes may not accurately reflect the current state of the features as observed in Sentinel imagery. For instance, the banks of a watercourse, delineated based on aerial imagery, may undergo changes over time, spanning a year. These changes can result from various factors such as natural processes or human activities, causing the banks to shift or erode. Consequently, the annotations based on older aerial imagery may not capture these temporal variations. 
    
\end{itemize}

Figure~\ref{fig:patches} gives an example of aerial patches, corresponding extracted super-patch (with the aerial patch footprint in black outlines) and annotation patches. The interest of the extended spatial information provided by the Sentinel-2 super-patches is particularly visible in the last two rows of Figure~\ref{fig:patches}. Indeed, the location on a beach or on a lake is difficult to determine from the aerial image alone, and could easily be confused with the sea for example in the last row.

\section*{\textbf{Semantic classes and frequency}}
\begin{figure}[!ht]
\centering\includegraphics[width=1\linewidth]{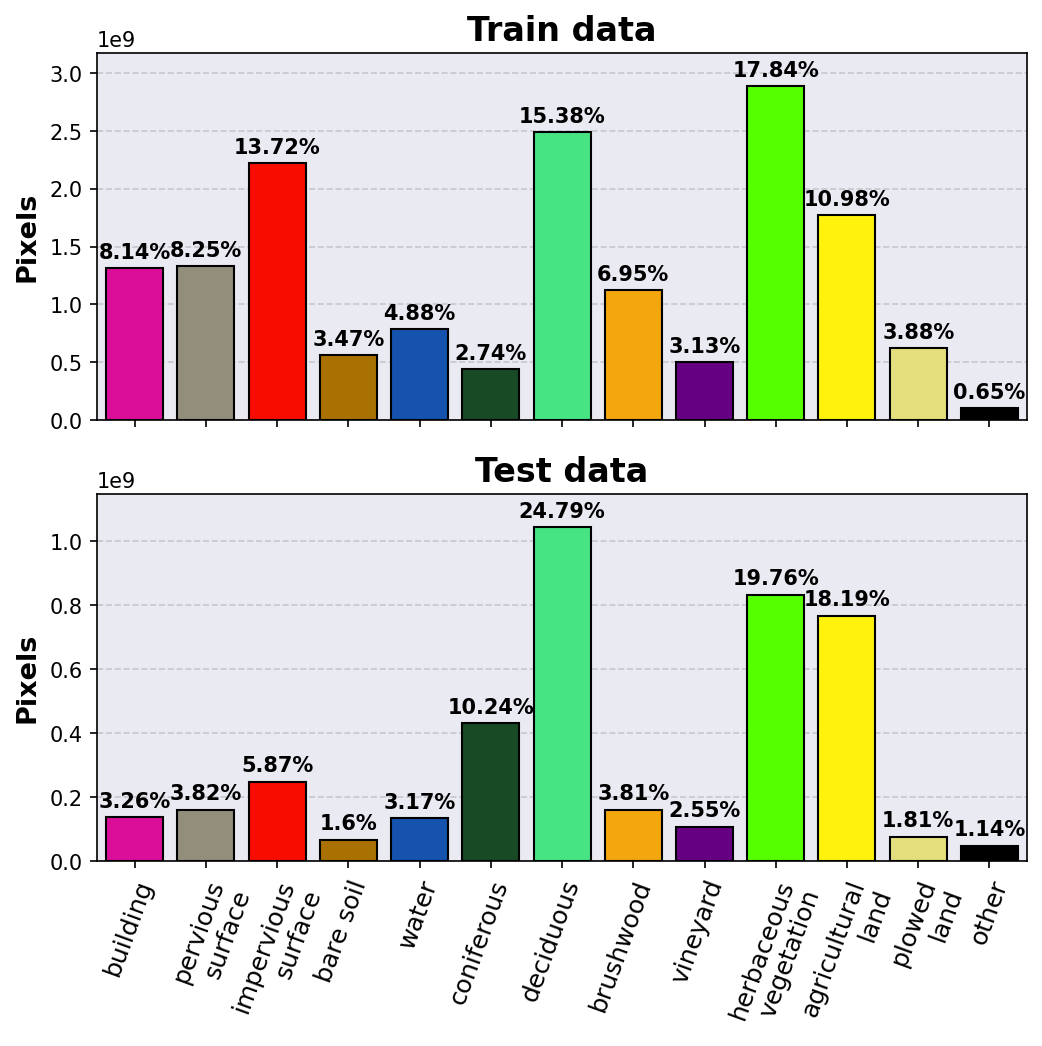}
\caption{Class distribution of the train dataset (\textit{top}) and test dataset (\textit{bottom}).}
\label{fig:class_freq_train}
\end{figure}

For a complete description of the 18 semantic classes (+1 \textit{other} class corresponding to unknown land cover) that are available in the annotations patches and information about the annotation process, see \cite{FLAIR-1}. As explained in \cite{FLAIR-1}, 5 from the 18 classes were merged into the \textit{other} class, due to strong under-representation ($<$ 1\% of the complete dataset). This results in a nomenclature of 12 classes plus the \textit{other} class. Table~\ref{tab:nomenclature} indicates the resulting classes, value in the MSK patches and frequency across the entire \mbox{FLAIR \#2} dataset. The class distribution in percentages of the train and test datasets are presented in Figure~\ref{fig:class_freq_train}.

\begin{table}[htpb]
\footnotesize
\centering
\setlength{\tabcolsep}{4.9pt}
\renewcommand{\arraystretch}{1.6}
\begin{tabular}{p{1cm}lccr}
& \textbf{Class}         & \textbf{MSK}           & \textbf{Pixels}           & \textbf{\%}                \\ \hline
\cellcolor[HTML]{db0e9a} & building               & 1                         & 1,453,245,093 & 7.13       \\
\cellcolor[HTML]{938e7b} & pervious surface       & 2                         & 1,495,168,513 & 7.33       \\
\cellcolor[HTML]{f80c00} & impervious surface     & 3                         & 2,467,133,374 & 12.1       \\
\cellcolor[HTML]{a97101} & bare soil              & 4                         & 629,187,886   & 3.09       \\
\cellcolor[HTML]{1553ae} & water                  & 5                         & 922,004,548   & 4.52       \\
\cellcolor[HTML]{194a26} & coniferous             & 6                         & 873,397,479   & 4.28       \\
\cellcolor[HTML]{46e483} & deciduous              & 7                         & 3,531,567,944 & 17.32      \\
\cellcolor[HTML]{f3a60d} & brushwood              & 8                         & 1,284,640,813 & 6.3        \\
\cellcolor[HTML]{660082} & vineyard               & 9                         & 612,965,642   & 3.01       \\
\cellcolor[HTML]{55ff00} & herbaceous vegetation  & 10                        & 3,717,682,095 & 18.24      \\
\cellcolor[HTML]{fff30d} & agricultural land      & 11                        & 2,541,274,397 & 12.47      \\
\cellcolor[HTML]{e4df7c} & plowed land            & 12                        & 703,518,642   & 3.45       \\
\cellcolor[HTML]{000000} & other                  & \textbf{\textgreater{}13} & 153,055,302   & 0.75       \\ \hline
\end{tabular}
\caption{Semantic classes of the main nomenclature of the \mbox{FLAIR \#2} dataset and their corresponding MSK values, frequency in pixels and percentage among the entire dataset.}
\label{tab:nomenclature}
\end{table}

The current test dataset has a different sampling than \mbox{FLAIR \#1}. The use of satellite time series to inject temporal information is especially relevant for natural surfaces with \textit{e.g.} a seasonal variation. Therefore, the classes of forests (coniferous and deciduous), agricultural land and herbaceous cover were favored, accounting for 72.98\% of the test dataset.

\begin{figure*}[!ht]
    \centering
    \includegraphics[width=1\linewidth]{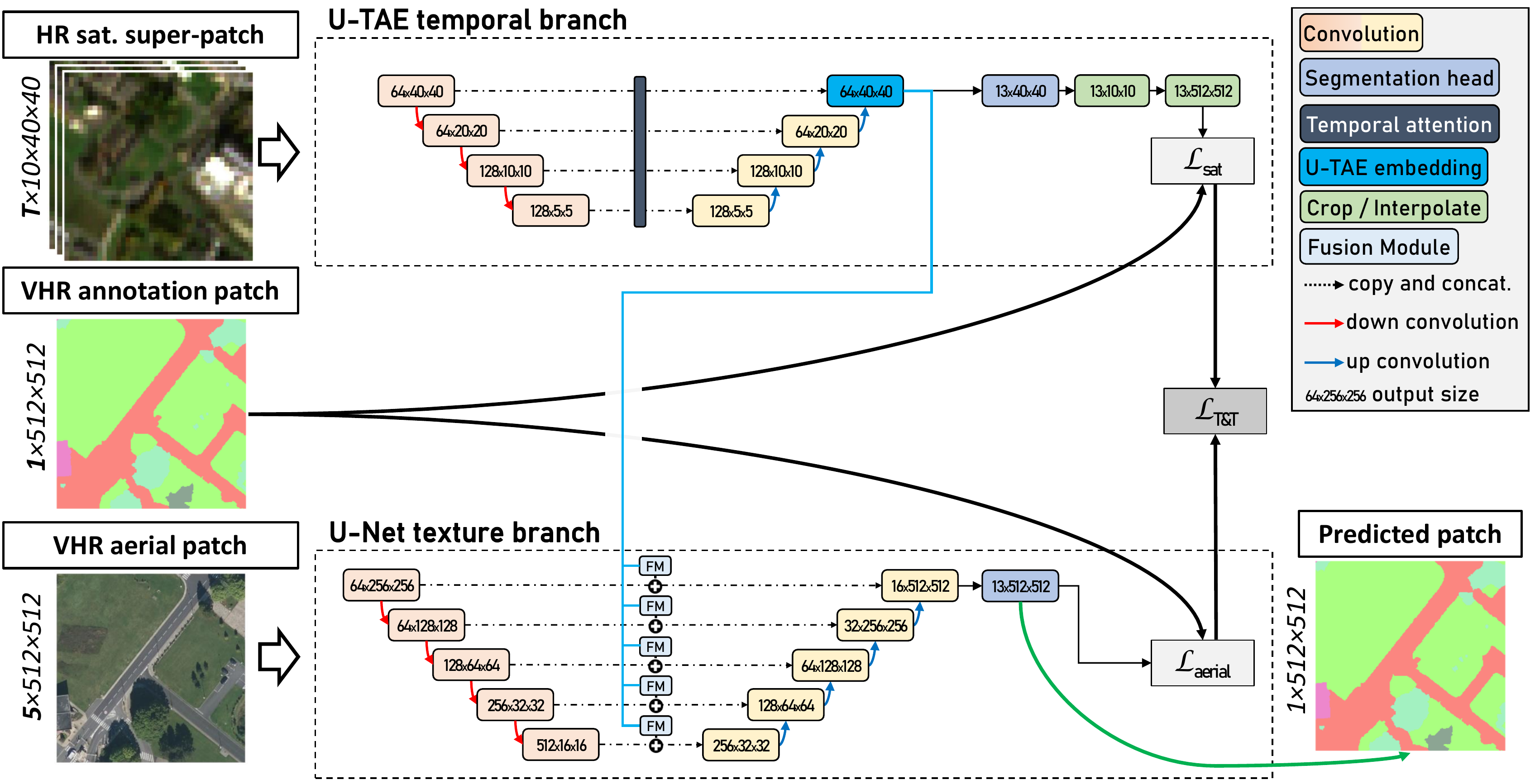}
    \caption{\textit{Texture and Time} extraction network including two branches: i) a \mbox{U-TAE} network applied to the Sentinel-2 super-patch time series and ii) a \mbox{U-Net} network applied to the mono-date aerial imagery patch. The last decoder layer yielded features from the \mbox{U-TAE} branch are used as embeddings added to the features of the \mbox{U-Net} branch, integrating temporal information from the time series and spatial information from the extended super-patch. The light-blue fusion type modules are enabled or not and varying according to the fusion method.}
    \label{fig:net}
\end{figure*}

\begin{figure}[!ht]
    \centering
    \includegraphics[width=0.97\linewidth]{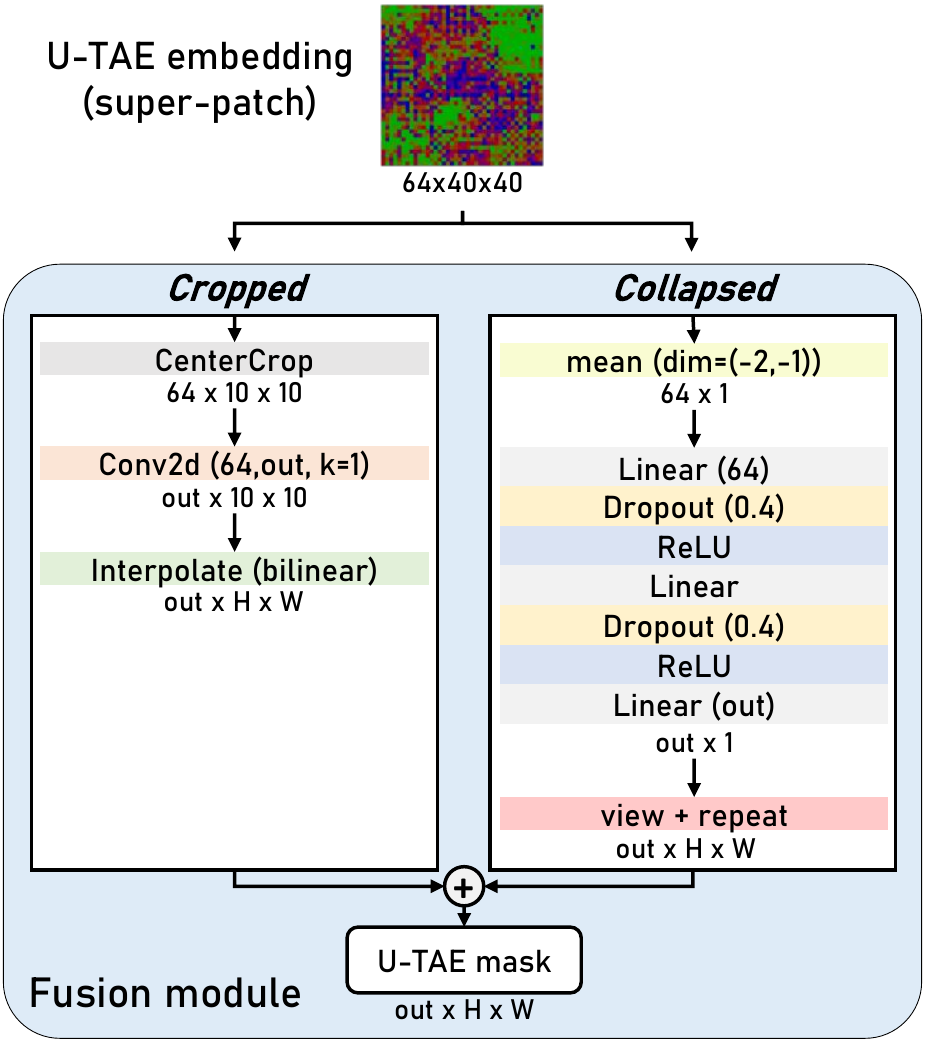}
    \caption{Fusion module taking as input the last \mbox{U-TAE} embeddings. This module is applied to each stages of the \mbox{U-Net} encoder feature maps. \textit{out} corresponds to the channel size of the \mbox{U-Net} encoder feature map and \textit{H} and \textit{W} to the corresponding spatial dimensions.}
    \label{fig:fusion}
\end{figure}

\section*{\textbf{Benchmark architecture}}
\textbf{Network definition}: to capture both spatial and temporal information from very high resolution aerial images and high-resolution satellite images, we propose a two-branch architecture called \textbf{\mbox{U-T\&T}}, for \textit{Textural} and \textit{Temporal} information. The model allows enables the fusion of learned time series-related information with the low-level representations of mono-date learned information. The \mbox{U-T\&T} model  combines two commonly used architectures: 

\begin{itemize}[topsep=8pt]
    \setlength\itemsep{1em}
    
    \item[$\blacktriangleright$] \textbf{\mbox{U-Net} (spatial/texture branch)}: to handle the aerial imagery patches, a \mbox{U-Net} architecture \cite{U-Net} is adopted. The encoder is using a ResNet34 backbone model \cite{ResNet} which has been pre-trained on the ImageNet dataset \cite{imagenet}. The \mbox{U-Net} branch has $\approx$ 24.4\:M parameters. Ith closely resembles to the model described in the \mbox{FLAIR \#1} datapaper \cite{FLAIR-1}, ensuring consistency and comparability with prior work.
    
    \item[$\blacktriangleright$] \textbf{\mbox{U-TAE} (spatio-temporal branch)}: a \mbox{U-TAE} \cite{UTAE} architecture focuses on extracting and incorporating both spatial and temporal information from the Sentinel-2 time series data. This architecture is based on \mbox{U-Net} but incorporates a Temporal self-Attention Encoder (TAE) component taking as input the lowest resolution features of the convolutional encoder to generate set of attention masks that capture the temporal dependencies within the time series data. These attention masks are then applied at all resolutions upon the decoding process, enabling the model to capture spatio-temporal patterns in the data. 
    
\end{itemize}

Figure~\ref{fig:net} provides an overview of the proposed method, which combines the \mbox{U-TAE} and \mbox{U-Net} architectures. The main idea behind this approach is to incorporate features learned by the \mbox{U-TAE} branch, which considers the temporal dimension and a wider spatial context, into the \mbox{U-Net} branch, which focuses on aerial imagery. However, a key constraint is the significant difference in spatial resolution between the satellite and aerial data. With the satellite imagery having a spatial resolution 50 times lower than the aerial imagery (10\:m versus 0.2\:m), early and late fusion strategies (\textit{i.e.}, fusion at input or prediction levels) are not viable due to the large size disparity. To address this, a \textit{Fusion Module} is introduced, depicted in Figure~\ref{fig:fusion}, which enables mid-stage fusion of features from both branches:

\begin{itemize}[topsep=5pt]
    \setlength\itemsep{1em}
    \item[$\blacktriangleright$] \textbf{Fusion Module}: the fusion module takes as input the \mbox{U-TAE} embedding (last feature maps of the \mbox{U-TAE} decoder, shown in blue in Figure~\ref{fig:net}) and is applied to each stage of the \mbox{U-Net} branch. Within the \textit{Fusion Module}, two sub-modules have different purposes and focus on distinct aspects:\\
    
    \begin{itemize}
        \setlength\itemsep{1em}
        
        \item \textbf{\textit{Cropped}}: this sub-module aims at incorporating information from the \mbox{U-TAE} super-patch embedding into the spatial extent of the aerial parches. The \mbox{U-TAE} embedding is first cropped to match the extent of the aerial patch. This cropped embedding is then fed to a single convolution layer, which produces a new channel dimension size that aligns with the one of the \mbox{U-Net} encoder feature maps channel size. The output of this convolutional layer is then passed through an interpolation layer that utilizes bilinear resampling. This interpolation ensures that the spatial dimensions matches those of the \mbox{U-Net} feature maps. 

        \item \textbf{\textit{Collapsed}}: this sub-module is designed to preserve spatial information from the extended super-patch, which will be integrated into the \mbox{U-Net} feature maps. Initially, the spatial dimension of the U-TAE is collapsed into a single value per channel, typically by taking the mean. The resulting vector is then fed into a shallow Multi-Layer Perceptron (MLP) consisting of three linear layers with dropout regularization and Rectified Linear Unit (ReLU) activation. The output size of the MLP is adjusted to match one of the \mbox{U-Net} encoder feature maps channel size.Subsequently, for each value in the obtained vector, the value is duplicated across the spatial dimension of the corresponding \mbox{U-Net} encoder feature maps.    
    \end{itemize}
    
    \item[] Both the \textit{cropped} and \textit{collapsed} sub-modules produce a mask of size \textit{out}$\times$\textit{H}$\times$\textit{W}, where \textit{out}, \textit{H}, and \textit{W} correspond to the targeted feature map dimensions of the \mbox{U-Net} model. These masks, generated separately, are initially added together to integrate spatio-temporal information from the Sentinel-2 satellite time series. The resulting combined mask is added to the feature maps of the \mbox{U-Net} model. This integration step allows the spatio-temporal information captured by the \textit{cropped} and \textit{collapsed} sub-modules from the Sentinel-2 satellite time series to be incorporated into the \mbox{U-Net's} feature representation.   
\end{itemize}

\vspace{1cm}
\textbf{Network supervision}: a single $\mathcal{L_{T\&T}}$ loss is used to monitor the training, which is the sum of two auxiliary losses $\mathcal{L}$$_{sat}$ and $\mathcal{L}$$_{aerial}$, obtained respectively from the \mbox{U-TAE} and \mbox{U-Net} branches. The two branches are using a categorical Cross Entropy (CE) cost-function, suitable for multi-class supervised classification task : 

\vspace{0.1cm}
\begin{equation*}
\mathcal{L_{CE}} = -\sum_{i=1}^nt_{i}\log(p_{i})\quad,
\end{equation*}
\vspace{0.05cm}
\begin{equation*}
\mathcal{L}_{T\&T} = \mathcal{L}_{CE\;aerial} + \mathcal{L}_{CE\;sat}
\end{equation*}
\vspace{0.1cm}

where $t_i$ is the MSK label and $p_i$ the Softmax probability of the $i^{th}$ class.\\ 

The MSK files in the \mbox{FLAIR \#2} dataset are provided at a spatial resolution of 0.2\:m. The output of the U-TAE branch corresponds to a super-patch, which lacks annotations for most of its parts. To address this, the U-TAE outputs are initially cropped to match the extent of the corresponding aerial patch. Subsequently, they are interpolated to fit the spatial dimensions of the MSK files (512$\times$512 pixels). This interpolation ensures compatibility before calculating the $\mathcal{L}$$_{sat}$ loss.

\section*{\textbf{Benchmark metric}}
The evaluation methodology for the semantic segmentation task follows the approach used in the \mbox{FLAIR \#1} challenge \cite{FLAIR-1}. Initially, confusion matrices are calculated per patch, and then aggregated across the test dataset to create a single confusion matrix. To assess the performance of each semantic class, the Intersection over Union (IoU) metric, also known as the Jaccard Index, is computed. The IoU is calculated using the formula:\\
$$IoU = \frac{|U \cap V|}{|U \cup V| } = \frac{TP}{TP + FP + FN}  $$
\noindent where U denotes the intersection, V the union, TP the true positives, FP the false positives and FN the false negatives.\\

The mean Intersection over Union (\textbf{mIoU}) is then determined by taking the average of the per-class IoU values. However, since the \textit{other} class is not well-defined and is equivalent to void, it is excluded from the IoU calculations. Consequently, the mIoU is computed as the average of the IoUs from the remaining 12 classes.

\section*{\textbf{Benchmark framework and settings}}
The baselines are calculated using the efficient \textit{PyTorch Lightning} framework \cite{PNING}. For the implementation of the U-Net model, the \textit{segmentation-models-pytorch} library \cite{SMP} is exploited, while the U-TAE network is obtained from \cite{UTAE}. The U-TAE parameters are kept at their default values (as provided in the GitHub implementation), except for  the encoder and decoder widths.\\

For the training process, the train dataset consists of 40 domains, out of which 32 are used for training the model, while the remaining 8 domains are used for validation. The optimization technique employed is the stochastic gradient descent (SGD) with a learning rate of 0.001. A reduction strategy is implemented with a patience value of 10, allowing for adaptive adjustments to the learning rate during training. The maximum number of epochs is set to 100, but to prevent overfitting and save computational resources, an early stopping method is utilized with a patience of 30 epochs. A batch size of 10 is used for the baselines.\\

To ensure reproducibility and consistent results, all randomness is controlled by fixing the seed using the \textit{seed\_everything} function from the PyTorch library, with the seed value set to 2022. Twelve NVIDIA Tesla V100 GPUs with 32 GB memory each, located on a High-Performance Computing (HPC) cluster, are used to speed up experiments. The distributed data parallel (ddp) strategy is employed to leverage these computational resources efficiently, allowing for parallel training across multiple GPUs.\\

In the context of the U-TAE and U-Net models, both of which utilize CE loss, per class weighting is employed. When assigning weights to the classes, the \textit{other} class is explicitly set to 0, indicating that it does not contribute to the loss calculation. The remaining classes are assigned a weight of 1. However, in the case of the U-TAE model, the \textit{plowed land} class is also assigned a weight of 0 for the \mbox{U-TAE} CE loss. This decision is made because the \textit{plowed land} class is specifically designed for mono-temporal data. The inclusion of time series data introduces ambiguity with agricultural land, and therefore, setting the weight of the \textit{plowed land} class to 0 helps to mitigate this confusion.\\ 

In addition to these general hyperparameters, there are several other parameters and strategies that have been or could be explored further:
\begin{itemize}[topsep=8pt]
    \setlength\itemsep{1em}
    \item[\ding{233}] the \textbf{size of super-patches} refers to the dimensions, in terms of pixels, of the patches that are cropped from the super-areas. Different sizes can be tested, allowing for experimentation with smaller or larger super-patch sizes. However, it is important to note that there is a limit of 110 pixels for edge patches. The choice of super-patch size has an impact on the spatial context provided to both the \mbox{U-TAE} and \mbox{U-Net} branches through the \textit{collapsed} fusion sub-module.\\~\\
    \textit{\textbf{Baselines}: } the number 40 has been empirically determined and set as the baseline for this specific parameter.
    
    \item[\ding{233}] with the exception of the \textit{other} and \textit{plowed land} classes, no specific distinction or weighting has been applied during training between the classes and the network branches. However, it is possible to introduce \textbf{per-class weights} for both the $\mathcal{L}$${sat}$ and $\mathcal{L}$${aerial}$ losses. These weights can be determined based on expert knowledge to encourage specialization of one branch or the other on certain classes. Another approach is to apply weights during the summation of both losses to obtain $\mathcal{L}$$_{T\&T}$.\\~\\
    \textit{\textbf{Baselines}: } the \textit{other} class is assigned a weight of 0 for both branches, and the \textit{plowed land} class is assigned a weight of 0 for the \mbox{U-TAE} branch. The remaining classes are assigned a weight of 1. Additionally, no weights are applied during the summation of the $\mathcal{L}$${sat}$ and $\mathcal{L}$${aerial}$ losses.

    \item[\ding{233}] to prevent overfitting of the \mbox{U-TAE} branch and enhance the learned aerial features, we incorporate a \textbf{modality dropout mechanism}. This involves generating a random single value for each batch. If the generated value exceeds a specified threshold, provided as an input parameter, the \mbox{U-TAE} modality is dropped out, and only the \mbox{U-Net} branch is used for that particular batch.\\~\\
    \textit{\textbf{Baselines}: } considering the coarse spatial resolution of Sentinel-2 data, we set the modality dropout threshold relatively high, at a value of 0.5. This ensures that a significant portion of the batches will exclusively utilize the \mbox{U-Net} branch, thereby emphasizing the importance of the aerial imagery.

    \item[\ding{233}] to address the potential impact of cloud or snow in the Sentinel-2 time series, two strategies are implemented using the provided masks files. The first strategy, called \textbf{filter clouds}, involves examining the probability of cloud occurrence in the masks. If the number of pixels above a certain probability threshold exceeds a specified percentage of all pixels in the image, that particular date is excluded from the training process. This helps to mitigate the influence of cloudy or snowy images on the training data. The second strategy, known as \textbf{monthly average}, is specifically implemented to alleviate potential challenges faced by the \mbox{U-TAE} branch due to a large number of dates in the time series. In this strategy, a monthly average is computed using cloudless dates. If no cloudless dates are available for a specific month, fewer than 12 images may be used as input to the U-TAE branch.\\~\\
    \textit{\textbf{Baselines}: } a probability threshold of 0.5 is employed for filtering clouds or snow in the masks. Additionally, to be considered for exclusion, the clouds or snow must cover at least 60\% of the super-patch.

    \item[\ding{233}] similar to the \mbox{FLAIR \#1} approach, \textbf{metadata associated with each aerial patch} are integrated into the model. These metadata are encoded using positional encoding or one-hot encoding techniques (see \cite{FLAIR-1}). The encoded metadata are then passed through a MLP before being added to each \mbox{U-Net} encoder feature map.\\~\\
    \textit{\textbf{Baselines}: } a positional encoding of size 32 is used specifically for encoding the geographical location information.

    \item[\ding{233}] \textbf{data augmentation techniques} usually prevent overfitting and help generalization capabilities of a network. Simple geometric transformations are applied during the training process. These transformations include vertical and horizontal flips as well as random rotations of 0, 90, 180, and 270 degrees. This approach aligns with the methodology used in the \mbox{FLAIR \#1} challenge.\\~\\
    \textit{\textbf{Baselines}: } a data augmentation probability of 0.5 is used.

\end{itemize}

\newcommand{\false}[0]{\textcolor{red}{\ding{55}}}
\newcommand{\true}[0]{\textcolor{green}{\ding{52}}}
\begin{table*}[!ht]
\footnotesize
\centering
\setlength{\tabcolsep}{8pt}
\renewcommand{\arraystretch}{1.7}
\begin{tabular}{lccccccccc}

& \textbf{INPUT} & \textbf{FILT.} & \textbf{AVG M.} & \textbf{M.DR} & \textbf{MTD} & \textbf{AUG} & \textbf{PARA.} & \textbf{EP.} & \textbf{mIoU}  \\\hline 
\rowcolor[HTML]{D6D5D4}\textbf{\mbox{U-Net}} &  aerial &  -  & - & - & \false & \false & 24.4 & 62 & \textbf{0.5467}$\pm$0.0009 \\\hline
\textit{+MTD} &  aerial & - & - & - & \true & \false & 24.4 & 59 & \textbf{0.5473}$\pm$0.0017 \\\hline
\textit{+MTD +AUG}&  aerial & - & - & - & \true & \true & 24.4 & 52 & \textbf{0.5517}$\pm$0.0013 \\\hline 
\rowcolor[HTML]{D6D5D4}\textbf{\mbox{U-T\&T}} &  aerial+sat & \false & \false & \false & \false & \false & 27.3 & 9 & \textbf{0.5490}$\pm$0.0072\\\hline
\textit{+FILT} &  aerial+sat & \true & \false & \false & \false & \false & 27.3 & 11 & \textbf{0.5517}$\pm$0.0135\\\hline
\textit{+AVG M} &  aerial+sat & \false & \true & \false & \false & \false & 27.3 & 10 & \textbf{0.5504}$\pm$0.0067\\\hline
\textit{+M DR} &  aerial+sat & \false & \false & \true & \false & \false & 27.3 & 27 & \textbf{0.5354}$\pm$0.0104\\\hline
\textit{+MTD} &  aerial+sat & \false & \false & \false & \true & \false & 27.3 & 7 & \textbf{0.5494}$\pm$0.0064\\\hline
\textit{+AUG} &  aerial+sat & \false & \false & \false & \false & \true & 27.3 & 22 & \textbf{0.5554}$\pm$0.0146\\\hline
\textit{+FILT +AVG M +M DR +MTD +AUG}  &  aerial+sat & \true & \true & \true & \true & \true & 27.3 & 36 & \textbf{0.5323}$\pm$0.0016\\\hline
\rowcolor[HTML]{ebfce9}\textit{+FILT +AVG M +AUG}  &  aerial+sat & \true & \true & \false & \false & \true & 27.3 & 26 & \textbf{0.5623}$\pm$0.0056\\\hline

\end{tabular}
\caption{Baseline results of ResNet34/\mbox{U-Net} architecture with aerial imagery only and \mbox{U-T\&T} with aerial and satellite imagery on the \mbox{FLAIR \#2} test set. Results are averages of 5 runs of each configuration. \textbf{FILT}: filter Sentinel-2 acquisition with masks (clouds \& snow); \textbf{AVG M.}: monthly average of all Sentinel-2 acquisitions; \textbf{M.DR}: modality dropout of the \mbox{U-TAE} branch; \textbf{MTD}: metadata for aerial imagery added; \textbf{AUG}: geometric data augmentation for aerial imagery; \textbf{PARA.}: number of parameters of the network; \textbf{EP.}: best validation loss epoch.}
\label{tab:baseline}
\end{table*}

\section*{\textbf{Benchmark results}}
Firstly, an evaluation is conducted on a \mbox{U-Net} model that incorporates only aerial imagery, resembling the approach used in the \mbox{FLAIR \#1} challenge. The evaluation involves assessing the model's performance using the code provided in the GitHub repository (accessible at \cite{FLAIR_page}). Following this, the results obtained from applying the two-branches \mbox{U-T\&T} model are reported. Additionally, various parameters and strategies mentioned earlier are tested.\\

The models used in the evaluation were trained using a consistent train/validation/test split and the parameters previously specified. The training dataset consisted of 61,712 aerial imagery patches, and for the \mbox{U-T\&T} approach, an additional 41,029 (unfiltered) Sentinel-2 acquisitions are included. During the inference phase, the models were applied to 16,050 patches of aerial imagery and 10,215 (unfiltered) satellite acquisitions from the test dataset. The reported results represent the average mIoU scores obtained from five separate runs of each model configuration. Additionally, the standard deviation of the mIoU scores across the five runs is provided, indicating the degree of variability in the performance of the models.\\   

The results obtained from the different experiments are presented in Table~\ref{tab:baseline}. When using only aerial imagery and a \mbox{U-Net} model, the highest mIoU score of 0.5517 is achieved by integrating aerial metadata and employing data augmentation techniques. In the case of jointly utilizing aerial and satellite imagery with the \mbox{U-T\&T} model, the baseline model yields a slightly better mIoU score compared to the aerial-only baseline (0.5490 versus 0.5467), but it also exhibits a higher standard deviation in the results.\\

Table~\ref{tab:baseline} also includes the results obtained when implementing additional strategies individually, as described in the Benchmark framework and settings section. It is observed that using modality dropout leads to a decrease in the mIoU score. Integrating aerial metadata into the \mbox{U-Net} branch only marginally improves the results. However, for the remaining three strategies, namely filtering the dates using cloud and snow masks, performing a monthly average of Sentinel-2 acquisitions, and applying data augmentation, the mIoU scores improve. By combining these three strategies, a mIoU score of 0.5623 is achieved, corresponding to a 2.85\% increase compared to the \mbox{U-Net} baseline.\\

\newcommand*\best{\cellcolor[HTML]{ebfce9}}
\begin{table*}[!ht]
\footnotesize
\centering
\setlength{\tabcolsep}{3pt}
\renewcommand{\arraystretch}{1.7}
\begin{tabular}{lccccccccccccc}
 & \makecell{mIoU\\ \:} & \makecell{building\\ \:} & \makecell{pervious\\surface} & \makecell{impervious\\surface}& \makecell{bare soil\\ \:} & \makecell{water \\ \:} & \makecell{coniferous\\ \:} & \makecell{deciduous\\ \:} & \makecell{brushwood\\ \:} & \makecell{vineyard\\ \:} & \makecell{herbaceous\\vegetation}& \makecell{agricultural\\land} & \makecell{plowed\\land}\\ \hline
\multicolumn{1}{|l|}{\cellcolor[HTML]{E1E1E1}\mbox{U-Net}} & 0.5470 & 0.8009 & 0.4727 & 0.6988 & 0.3076 & 0.7985 & 0.5758 & 0.7014 & 0.2392 & 0.6012 & 0.4653 & 0.5449 & \multicolumn{1}{c|}{\textbf{\best 0.3583}} \\
\multicolumn{1}{|l|}{\cellcolor[HTML]{E1E1E1}\mbox{U-T\&T}} & 0.5594 & 0.8285 & 0.4980 & 0.7204 & 0.2982 & \textbf{\best 0.8009} & 0.6041 & 0.7189 & \textbf{\best 0.2541} & 0.6580 & 0.4684 & 0.5478 & \multicolumn{1}{c|}{0.3157} \\
\multicolumn{1}{|l|}{\cellcolor[HTML]{E1E1E1}\mbox{U-T\&T} best} & \textbf{\best 0.5758} & \textbf{\best 0.8368} & \textbf{\best 0.4995} & \textbf{\best 0.7446} & \textbf{\best 0.3959} & 0.7952 & \textbf{\best 0.6339} & \textbf{\best 0.7239} & 0.2485 & \textbf{\best 0.6678} & \textbf{\best 0.4750} & \textbf{\best 0.5513} & \multicolumn{1}{c|}{0.3381} \\ \hline
\end{tabular}
\caption{Per-class IoU results of the \mbox{U-Net} baseline (aerial imagery), the \mbox{U-T\&T} baseline (aerial and satellite imagery) and the best \mbox{U-T\&T} configuration.}
\label{tab:baseline_ious}
\end{table*}

The per-class IoU scores for three models are provided in Table~\ref{tab:baseline_ious}. The three models considered are the \mbox{U-Net} baseline, the \mbox{U-T\&T} baseline, and the \mbox{U-T\&T} model with dates filtering of Sentinel-2, monthly average, and data augmentation. These models were selected based on achieving the highest mIoU scores among the five runs. Among the 12 classes, the \mbox{U-Net} baseline outperforms the other models by having a higher IoU score only for the \textit{plowed land} class, with a marginal improvement of 0.02 points compared to the \mbox{U-T\&T} best model. On the other hand, the \mbox{U-T\&T} baseline model performs better in predicting the \textit{water} and \textit{brushwood} classes, but the differences in IoU scores are quite close to the other models. For the remaining nine classes, the \mbox{U-T\&T} best model surpasses the other models, exhibiting notable improvements in classes such as \textit{buildings}, \textit{impervious surfaces}, \textit{bare soil}, \textit{coniferous}, and \textit{vineyards}. These improvements highlight the effectiveness of the \mbox{U-T\&T} model with the integrated strategies of dates filtering, monthly average, and data augmentation.\\

\begin{figure}[!ht]
    \centering
    \includegraphics[width=0.97\linewidth]{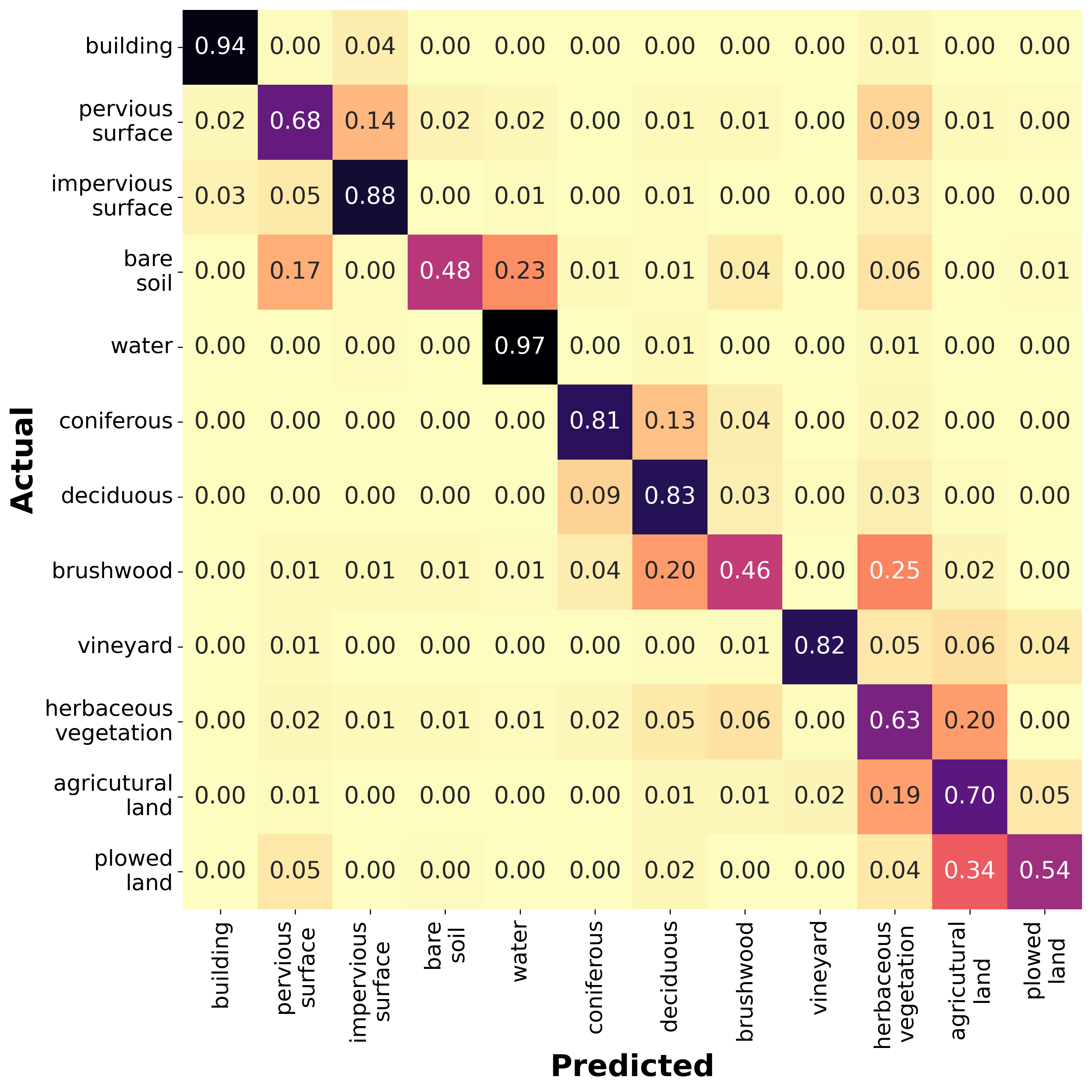}
    \caption{\mbox{U-T\&T} best model confusion matrix of the test dataset normalized by rows.}
    \label{fig:confmat}
\end{figure}

Figure~\ref{fig:confmat} illustrates the confusion matrix of the best \mbox{U-T\&T} model. This confusion matrix is derived by combining all individual confusion matrices per patch and is normalized by rows. The analysis of the confusion matrix shows that the best \mbox{U-T\&T} model achieves accurate predictions with minimal confusion in the majority of classes. However, when it comes to natural areas such as \textit{bare soil} and \textit{brushwood}, although there is improvement due to the use of Sentinel-2 time series data, a certain level of uncertainty remains. These classes exhibit some confusion with semantically similar classes, indicating the challenge of accurately distinguishing them.

\begin{figure*}[!ht]
    \centering
    \captionsetup{justification=centering,margin=2cm,}
    \includegraphics[width=0.90\linewidth]{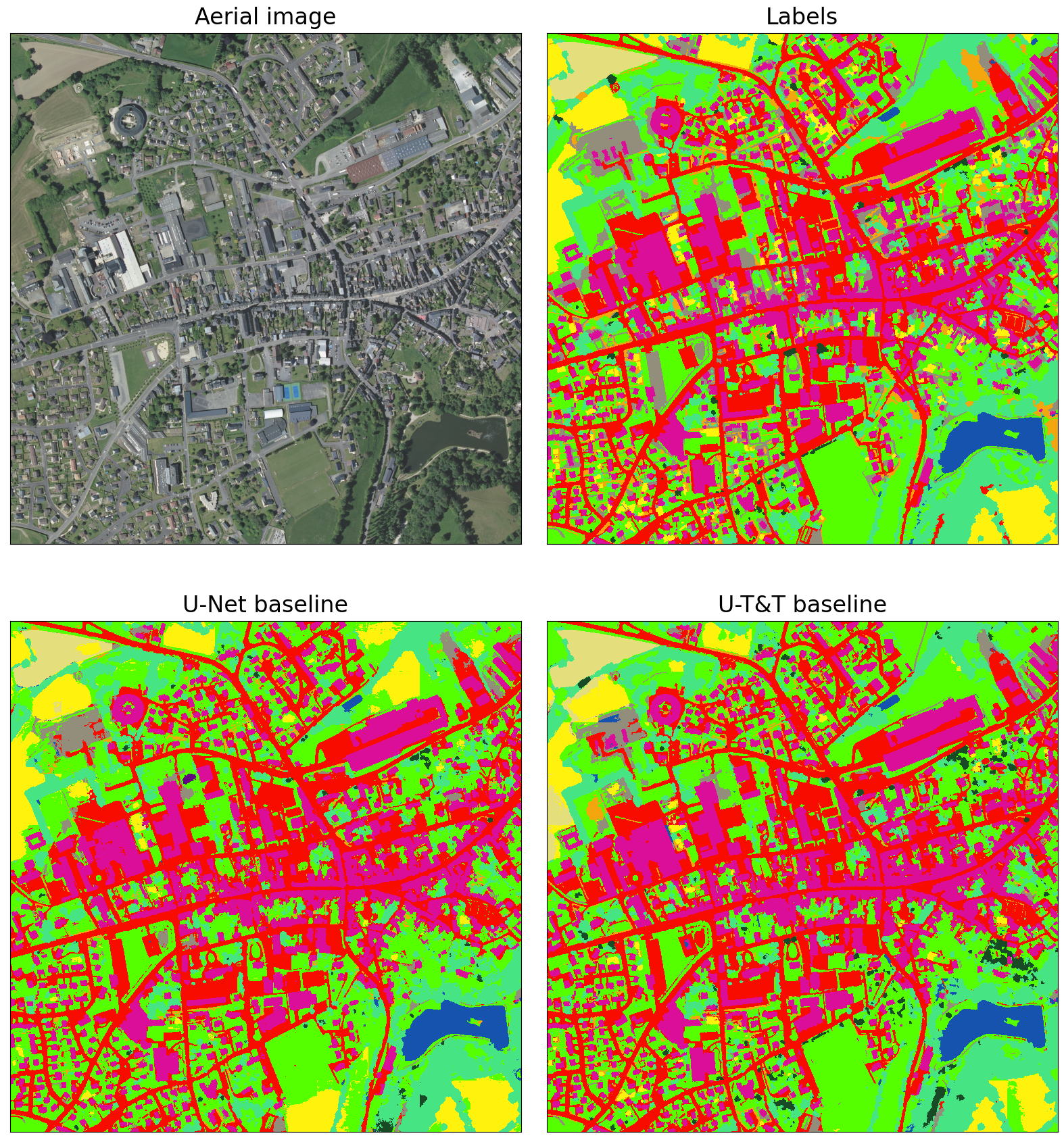}
    \caption{Comparison of results obtained on an area of the test dataset.\\\textit{Top row:} very high spatial resolution aerial imagery and corresponding labels.\\\textit{Bottom row:} \mbox{U-Net} baseline from aerial imagery and \mbox{U-T\&T} baseline from aerial and satellite imagery.}
    \label{fig:baselines_comp}
\end{figure*}

Figure~\ref{fig:baselines_comp} showcases an example that illustrates the results of both the \mbox{U-net} baseline and\mbox{U-T\&T} baseline models in relation to the aerial imagery and the corresponding annotations.

\section*{\textbf{Acknowledgment}}
The experiments conducted in this study were performed using HPC/AI resources provided by GENCI-IDRIS (Grant 2022-A0131013803). This work was supported by the European Union through the project "Copernicus / FPCUP," as well as by the French Space Agency (CNES) and Connect by CNES. The authors would like to acknowledge the valuable support and resources provided by these organizations.

\section*{\textbf{Data access}}
The dataset and codes used in this study will be made available after the completion of the \mbox{FLAIR \#2} challenge at the following website: \url{https://ignf.github.io/FLAIR/}.

\bibliographystyle{unsrt}
\bibliography{BIB.bib}

\clearpage

\end{document}